\documentclass{article}
\usepackage{ijcai22}

\usepackage{times}

\usepackage{soul}
\usepackage{url}
\usepackage[hidelinks]{hyperref}
\usepackage[utf8]{inputenc}
\usepackage[small]{caption}
\usepackage{graphicx}
\usepackage{amsmath}
\usepackage{booktabs}
\usepackage{multirow}
\usepackage{caption}
\usepackage{float}
\usepackage[toc,page]{appendix}
\usepackage[numbers]{natbib}
\usepackage{tabularx}
\usepackage[utf8]{inputenc}
\usepackage{amsthm}
\DeclareUnicodeCharacter{2212}{-}
\DeclareUnicodeCharacter{0301}{'}
\usepackage{acronym}
\usepackage{amsmath}
\usepackage{amssymb}

\acrodef{LIME}[LIME]{Local Interpretable Model-agnostic Explanations}
\acrodef{DL}[DL]{Deep-Learning}
\acrodef{ML}[ML]{Machine Learning}
\acrodef{XAI}[XAI]{eXplainable Artificial Intelligence}
\acrodef{SHAP}[SHAP]{SHapley Additive exPlanations}
\acrodef{Grad-CAM}[Grad-CAM]{Gradient-weighted Class Activation Mapping}
\acrodef{CNN}[CNN]{Convolutional Neural Network}
\acrodef{AI}[AI]{Artificial Intelligence}

\title{Fruit Ripeness Classification: a Survey}

\author{
Matteo Rizzo$^1$\footnote{Contact Author}\and
Matteo Marcuzzo$^{1}$\and
Alessandro Zangari$^1$\and
Andrea Gasparetto$^1$\and
Andrea Albarelli$^1$\and
\affiliations
$^1$ Ca' Foscari University
\emails
\{matteo.rizzo, matteo.marcuzzo, alessandro.zangari, andrea.gasparetto, albarelli\}@unive.it
}

\begin{document}

\maketitle

\begin{abstract}
Fruit is a key crop in worldwide agriculture feeding millions of people. The standard supply chain of fruit products involves quality checks to guarantee freshness, taste, and, most of all, safety. An important factor that determines fruit quality is its stage of ripening. This is usually manually classified by field experts, making it a labor-intensive and error-prone process. Thus, there is an arising need for automation in fruit ripeness classification. Many automatic methods have been proposed that employ a variety of feature descriptors for the food item to be graded. Machine learning and deep learning techniques dominate the top-performing methods. Furthermore, deep learning can operate on raw data and thus relieve the users from having to compute complex engineered features, which are often crop-specific. In this survey, we review the latest methods proposed in the literature to automatize fruit ripeness classification, highlighting the most common feature descriptors they operate on.
\end{abstract}

\section{Introduction}
\label{sec:introduction}

Fruit constitutes a commercially important and nutritionally indispensable food commodity. It has a fundamental role as a part of a balanced diet by supplying necessary growth-regulating factors that are essential for ensuring good health \citep{references_1_v_prasanna_fruit_2007}. To meet nutritional and market quality standards, it is crucial for fruit to be delivered at a proper ripeness stage. Moreover, grading fruit crops based on ripeness enables harvest optimization and all the subsequent links in the food supply chain. However, the classification of fresh fruit according to its ripeness is typically a subjective and tedious task. This is traditionally based on human effort, which makes it sub-optimal. Consequently, there is a growing interest in developing techniques to automate the fruit grading process. Machine Learning (ML) methods such as Support Vector Machines (SVMs), decision trees, and K-Nearest Neighbour (KNN) algorithms have been successfully applied to classification problems in the literature, particularly for fruit classification. More recently, Deep Learning (DL) methods such as Artificial Neural Networks (ANNs) and one of their derivations, Convolutional Neural Networks (CNNs), have also been applied to the fruit classification class showing very promising results.

\subsection{Challenges Of The Task}

The particularities of each ripeness classification problem make it difficult, if not impossible, to select a general technique that applies to all types of fruit. In fact, there is an incredibly wide range of fruit varieties in nature, and this gives birth to a complex taxonomy. Each variety of fruit may differ from the others in shape, color, texture, and many other features. Sometimes this difference is apparent (\textit{e.g.}, strawberries vs pineapple), while sometimes distinguishing two fruit classes is non-trivial even for experts (\textit{e.g.}, apples vs Asian pears). Moreover, each fruit type may embed highly variable features. This is the case, for example, of apples, where a Fuji apple has a completely different color and texture with respect to a Granny Smith apple (despite sharing a similar shape). Thus, the upstream task of fruit recognition and the downstream task of ripeness classification are both challenging. On the other hand, it is essential to enable automation within the supply chain of fruit crops to keep up with the worldwide demand for high-quality, safe food. The following sections depict the challenges and opportunities of fruit ripeness classification, paving the way to the statement of the contributions of this survey. 

\subsection{Ripeness Standards}

As mentioned, on top of mere discrimination between types of fruit, it is of utmost importance to ensure food quality. This is both for safety reasons and because of the practical fact that high-quality food is more appealing to the market. One core factor that influences the economic value of fruit is the relatively short ripening period and reduced post-harvest life. Large amounts of fruit are kept for a notably long time in storage for transportation and during this period it continues to mature. Thus, ripeness is a crucial property of fruit along the supply chain. Fruit ripening is a highly coordinated, genetically programmed, and irreversible phenomenon involving a series of physiological, biochemical, and organoleptic changes, that ultimately lead to the development of soft, edible, and ripe fruit with desirable quality attributes \citep{references_1_v_prasanna_fruit_2007}. On the producers' side, excessive textural softening during ripening may lead to problematic effects upon storage, including dents and poor shaping. On the consumers' side, the appearance of the product is one of the most worrying issues for producers, as it has a strong influence on product quality and consumers' preferences. However, up to this day, optimal harvest dates and prediction of storage life are still mainly based on subjective interpretation and practical experience \citep{mendoza_application_2004}. Accordingly, an objective and accurate ripeness assessment of crops is important in ensuring the optimum yield of quality products. The quality of the fruit (as measured by aroma, flavor, color, and textural characteristics) constantly changes during fruit development from pre-harvest through post-harvest stages as the fruit grows and ripens, as well as during maintenance in storage.

\subsection{Economical Impact Of Fruit Ripeness Classification}

It is imperative to improve the status of global food quality and security to meet the needs of the ever-enlarging world’s population. Increased fruit production through adding to the area cropped is not sustainable as land is a limited resource, thus productivity per unit of a land area must be increased. At the same time, there is a need to prevent waste. For fruit production, the timing of harvest is crucial to ensuring that the yield meets the commercial ripeness specifications. Over- or under-ripe fruit has a lower or even no retail value and represents significant income loss and a waste of resources. For the consumer, harvesting too early reduces the taste and quality of fruit whilst a late harvest can lead to reduced shelf life, poor appearance, and undesired flavors and odors.

\subsection{Predicting Fruit Ripeness}

Many computational methods are known to be effective in predicting fruit ripeness. Some of these require feature engineering (\textit{i.e.}, statistical and traditional ML algorithms) while others process raw data (\textit{i.e.}, DL algorithms). Each method is based on a specific feature representation of the fruit item to be graded. To collect discriminatory features, some methods require cheap and simple sensors (\textit{e.g.}, a consumer camera), and some methods require special (and usually expensive) sensors (\textit{e.g.}, an acoustic vibration detector). Similarly, some methods are item-destructive while others are non-destructive. In other words, the former type destroys the fruit item while assessing ripeness while the latter can still assess ripeness without any waste. In fact, the efficiency of destructive methods is limited, as they waste a part of the crop. Non-destructive on-plant assessment of fruit ripeness is generally preferable and has received increasing interest as it provides several advantages as compared to traditional destructive methods.

The phenotypic changes during fruit ripening are complex. In most cases, a green, hard, and immature fruit becomes more colorful, softer, sweeter, and aromatic. Numerous physical and chemical attributes can be quantified during ripening. These include size, shape, texture, firmness, external color, internal color, the concentration of chlorophyll, Soluble Solids Content (SSC), starch, sugars, acids, oils, and internal ethylene concentration \citep{references_1_v_prasanna_fruit_2007}. These features can be used to build non-destructive and efficient fruit ripeness classifiers. Although it is not realistic to simultaneously assess all the quality attributes in the field with non-destructive methods, destructive laboratory measurements are time-consuming due to the large number of samples needed to account for the within-field variability \citep{li_advances_2018}. Simple representative non-destructive measurements are thus required to assess the ripeness of fruit. A variety of non-destructive techniques and tools for features collection has been developed, mainly colorimeters,  Visible and Near InfraRed (VNIR) spectroscopy \citep{walsh_visible-nir_2020}, hyperspectral imaging \citep{su_application_2021}, visible imaging \citep{bulanon_visible_2011}, multispectral imaging \citep{shiddiq_wavelength_2022}, fluorescence imaging \citep{matveyeva_using_2022}, and electronic noses \citep{baietto_electronic-nose_2015}.

\begin{figure}[h]
    \centering
    \includegraphics[width=0.49\textwidth]{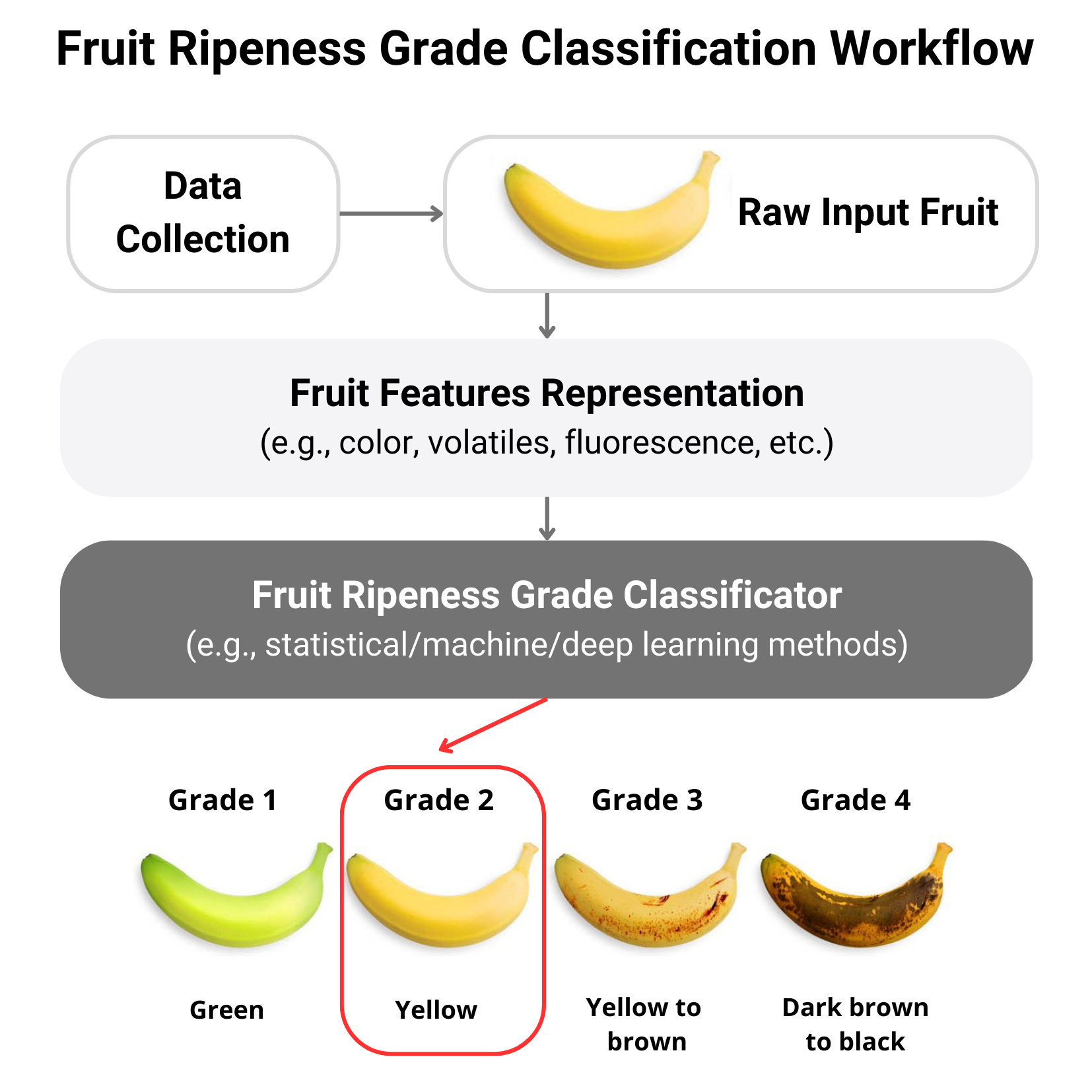}
    \caption{Example of workflow from the collection of raw data to the classification of ripeness grade for bananas.}
   \label{fig:workflow}
\end{figure}

In terms of methods used for coping with the ripeness classification task, seminal studies employed statistical techniques for sets of engineered features (\textit{e.g.}, \cite{mendoza_application_2004,jaradat_quality_2004,mangas_characterization_1998,nagata_estimation_2004}). More recently, ML techniques pushed the state-of-the-art. Simple baseline algorithms such as SVMs were applied to some of the above-mentioned feature sets and paved the way for a data-driven automatic classification of ripeness. Even more recently, DL algorithms have been explored for replacing traditional ML techniques and proved to be highly efficient and effective when very large datasets were available. The shift to a DL paradigm allows not to compute or collect engineered features, which is usually a time-consuming and error-prone task. Fig \ref{fig:workflow} summarizes the workflow from the data collection to the ripeness degree classification of some featurized representation of the raw input. 

\subsection{Contributions And Structure Of The Article}

This survey offers a high-level overview of the fruit ripeness classification task and of the methods that have been applied so far to reach excellent performance, with a major focus on the most novel DL techniques. We found two other surveys that have been dealing with fruit ripeness classification in the past. The work by \cite{randhawa_survey_2014} was published in 2014, thus it does not incorporate the most advanced DL methods. On the other hand, \cite{wankhade_survey_2021} focus mostly on image processing while our survey investigates not only state-of-the-art image processing techniques but also multiple feature descriptors for ripeness classification.

The rest of the text is organized as follows: §\ref{sec:preliminaries} provides a formal statement of the fruit ripeness classification problem together with basic biology notions on the subject of fruit ripening. §\ref{sec:feature-representations} focuses on the different feature representations that can be used (and possibly combined) to generate a feature description of a fruit item. §\ref{sec:data-preprocessing} briefly outlines the data preprocessing needed to apply classifiers to fruit ripening. §\ref{sec:approaches-to-classification} describes the collection of scientific literature that gathers the technical and theoretical effort to tackle the problem of fruit ripeness classification. §\ref{sec:optimal-harvest} embarks in the uncharted land of going over the specific problem (\textit{i.e.}, fruit ripeness classification) to the more general one (\textit{i.e.}, optimal harvest time estimation); §\ref{sec:perspectives} seizes the gaps in the above-mentioned literature and offers ideas on new research topics in the area; §\ref{sec:conclusions} concludes our work by synthesizing the contributions and themes treated in this paper.

\section{Preliminaries}
\label{sec:preliminaries}

This section introduces the formal statement of the fruit ripeness stage classification problem. Furthermore, basic biological notions enabling fruit ripening are discussed. These allow for a full understanding of some of the features and the representations used to grade fruit ripeness.

\subsection{Problem Statement}
\label{sec:problem-statement}

Given some dataset $X$ containing $|X|$ fruit appearance descriptors $x$ (\textit{e.g}., color, shape) for one variety of fruit, the ripeness classification problem requires learning a function $F$ that maps each item $x \in X$ to a predicted class $\hat{c}$, which is an approximation of the proper ripeness class $c \in C$. It is desirable that $\hat{c}=c$ holds. In other words, the learned function must map each input fruit description to its the chosen representation of its ground truth ripeness. The distance between the predicted class $\hat{c}$ and the ground truth $c$ is based on some notion of error (\textit{e.g.}, mean square error). 

Ground truth ripeness is generally assessed visually by a human operator led by some guidelines that describe the ripening stages (\textit{e.g.}, comparing the color of the peel of the fruit to standardized color charts). Human visual inspection is a highly subjective, tedious, time-consuming, and labor-intensive process. Tools-based techniques such as colorimeters or spectrometers, on the other hand, allow for accurate and reproducible measurements of the corresponding features with minimum influence by the observer or their surroundings \citep{anzalone_open-source_2013,das_ultra-portable_2016}.

\begin{figure}[b]
    \centering
    \includegraphics[width=0.49\textwidth]{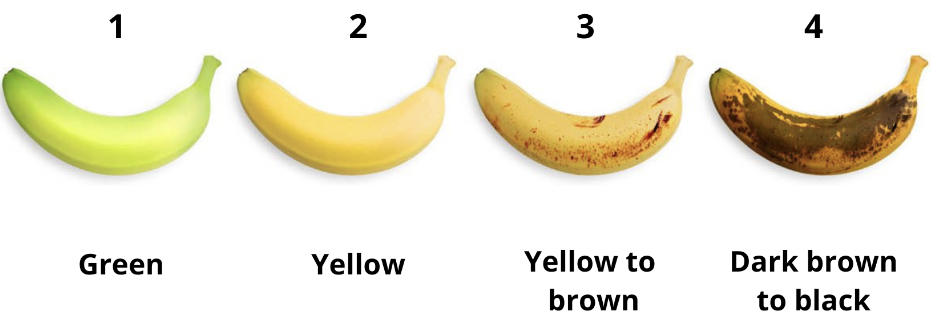}
    \caption{Example stages of ripeness of bananas.}
   \label{fig:bananas-ripeness}
\end{figure}

\begin{figure}[t]
    \centering
    \includegraphics[width=0.49\textwidth]{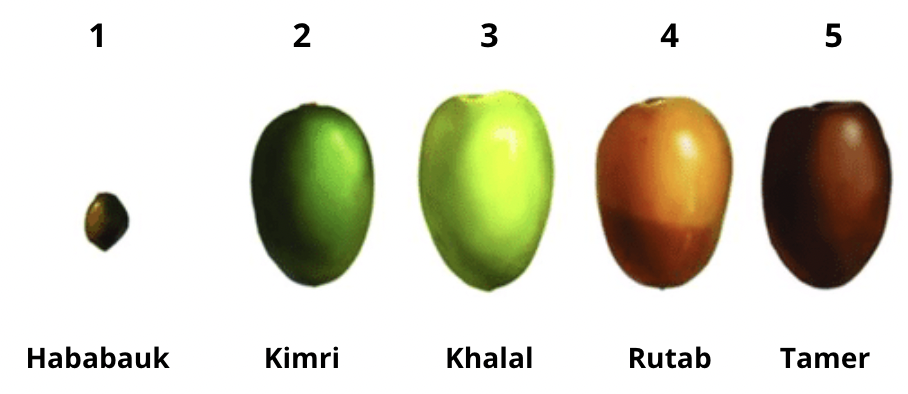}
    \caption{Example stages of ripeness of dates.}
    \label{fig:dates-ripeness}
\end{figure}

The fruit ripeness classification problem is wide-ranged, meaning it can differ dramatically based on the type of fruit being considered. First, $|C|$ can be arbitrarily large (although discrete) and vary based on the type of fruit. For instance, the ripeness of bananas can be categorized into four stages of ripening by visual inspection, which relates to pigment changes in the peel of the banana (\textit{i.e.}, Stage 1: Ripe, Stage 2: Partially ripe, Stage 3: Ripen and Stage 4: Over-ripe \citep{mazen_ripeness_2019}). This is exemplified in Fig \ref{fig:bananas-ripeness}. Similarly, the ripeness of dates can be categorized into five stages: Hanabauk, Kimri, Khalal, Rutab, and Tamer (see Fig \ref{fig:dates-ripeness}). Second, the variance between features within a single class can be very high, meaning that $X$ can be largely heterogeneous.

\subsection{Biology Of Fruit Ripening}
\label{sec:biology}

The development of fruit is characterized by a short period of cell division followed by a longer period of cell elongation by water uptake. The final fruit size mainly depends on the initial number of cells rather than the cell size \citep{references_1_v_prasanna_fruit_2007}. The process of ripening, on the other hand, is characterized by the development of peculiar color, flavor, texture, and aroma.

\subsubsection{The Role Of Ethylene}
Indisputable evidence indicates that ethylene plays a crucial role in the induction of ripening in fruit \citep{references_1_v_prasanna_fruit_2007}. Therefore, a main concern of the commercial post-harvest horticultural industry is to control the exposure of harvested fruit to ethylene. An aspect of fruit ripening that attracts molecular biologists is the prospect that ethylene induces a mature, non-growing tissue to rapidly differentiate into a new state, that is to switch from non-ripening to ripening. Because the tissue is already mature and the form and structure of cells decided, it is thought that the molecular steps involved when the switch in maturity is thrown may be relatively few. The ripening system, therefore, is studied for the subtle interactions between the signaling chemical ethylene and the fruit genome \citep{references_1_v_prasanna_fruit_2007}.

\subsubsection{Climacteric Vs Non-Climacteric Fruit}
Fruit can either be harvested at complete maturity or way before, especially when it is exported with long storage times. Each item is self-sufficient with its catalytic machinery to maintain an independent life, even when detached from the parent plant. Based on their respiratory pattern and ethylene biosynthesis during ripening, harvested fruit can be classified into two groups: (i) \emph{climacteric} and (ii) \emph{non-climacteric}. \emph{Climacteric} fruit is harvested at full maturity and is capable of maintaining the ripening process even when detached from the plant. The respiration rate and ethylene formation in this kind of fruit, though minimal at maturity, rise dramatically to a climacteric peak (\textit{i.e.}, the peak of edible ripeness), at the onset of ripening, after which it declines \citep{speirs_endopolygalacturonase_1990}. On the other hand, \emph{non-climacteric} fruit is not capable of continuing its ripening process once it is detached from the parent plant. Also, this type of fruit produces a very small quantity of endogenous ethylene and does not respond to external ethylene treatment.

\subsubsection{Texture And Softness}
Fruit ripening is associated with textural alterations. Textural change is the major event in fruit softening and an integral part of ripening \cite{gowda_studies_2001}. This process appears dramatically in climacteric fruits. Similarly, the softening process is an integral part of the ripening of almost all fruit. It has immense commercial importance because the post-harvest life of the fruit is to a large extent limited by increasing softness, which brings with it an increase in physical damage during handling and an increase in disease susceptibility.

\subsubsection{Fruit Ripening In The Literature}
Fruit ripening has been reviewed numerous times in the literature. In 1984, \cite{yang_ethylene_1984} discussed several aspects of the physiology and biochemistry of ripening, including the influence of exposition to ethylene. Some years later, in the study by \cite{speirs_endopolygalacturonase_1990}, some evidence seemed to show that protein and perhaps RNA synthesis played a role in the induction of ripening. The techniques of modern molecular biologists have enabled this theory to be examined in a precise way, and evidence for the direct genomics control of the ripening of climacteric fruits is now accumulating rapidly. Nonetheless, there is evidence that cellular compartments are modified through ripening, and recent proof suggests that lipid oxidation and/or phase changes within membranes contribute to changing metabolite distribution within cells as ripening proceeds. How novel transcriptional and/or translational events and changes in metabolite partitioning interact and contribute interdependently or independently to ripening is still a matter of conjecture. It may well be that both theories are more or less correct and both mechanisms are involved but to different extents in different types of fruit. Regardless, fruit ripening appears to be a well-regulated, genetically determined event. Coming as it does at the end of the development of an organism in an energy-rich tissue that is destined for spreading, it would be surprising if its mechanisms of control were the same for all fruit types.

\section{Fruit Feature Representations}
\label{sec:feature-representations}

When feeding data to a model (be it statistical, ML- or DL-based) a proper representation of the data involved is needed. More specifically, a set of features is collected to build a description of each data item. The more these features are representative and diversified among different classes, the more effective the classification. The next sections introduce the main types of features that are used in the literature to address the fruit ripeness classification problem.

\subsection{Colour}
\label{sec:colour}

Color is one of the first elements taken into consideration by producers and consumers when performing a qualitative assessment of a fruit's quality \citep{opara_assessment_2007}. This criterion is related to physical and chemical changes occurring during fruit ripening \citep{speirs_endopolygalacturonase_1990}. In many types of fruit, color changes during ripening occur due to chlorophyll degradation and the increase in the concentration of pigments such as carotenoids or polyphenols \citep{choo_fruit_2019}. Several fruit varieties have been studied for the relationship between maturity and color. These include: tomatoes \citep{shewfelt_prediction_1988}, oranges \citep{olmo_nondestructive_2000}, guavas \citep{mercado-silva_fruit_1998} and many more. To measure the changes in fruit color, the two major methods involve the usage of \emph{use of colorimeters} and \emph{image capture and analysis}.

\subsubsection{Colorimeters}
\textit{Colorimeters} are traditional non-destructive instruments used extensively in the fruit industry to measure fruit color \citep{hobson_assessing_1983}. They are more precise than human visual assessment and are able to adhere to common standards utilizing, for instance, the CIELAB color space. Portable colorimeters are now available commercially \citep{jha_modeling_2007} and can be carried to the field for \textit{in situ} data collection. Data obtained with colorimeters have been successfully correlated with fruit ripeness by using multivariate analysis (\textit{e.g.}, \cite{jha_modeling_2007}). However, the per-item measurement performed  by colorimeters limits the application of this tool for representing the ripeness of fruit in a whole crop. Moreover, the main disadvantage of colorimeters is that the surface color of the processed fruit item must be relatively uniform for the measurement to be meaningful. Otherwise, many spots on the sample must be measured to obtain a representative color profile, which becomes a labor-intensive job for the producer.

\subsubsection{Colour-imaging}

Ultimately, for some applications, colorimeters are simply not able to obtain representative color values due to the limited sampling area compared to the size of the fruit \citep{yam_simple_2004}. This limitation can be overcome by utilizing \textit{bi-dimensional color imaging}, which converts photons reflected from fruit skin to electrical signals, which are then received by a camera with a charge-coupled device or complementary metal oxide semiconductor sensors. Compared with a colorimeter, this type of color information can not only be obtained rapidly but also cover a larger area. Moreover, the equipment is flexible enough to be easily attached to moving platforms (\textit{e.g.}, tractors, robots, drones) for a rapid collection of multiple data measurements. However, even bi-dimensional imaging poses some challenges. First, it is difficult to segment fruit in an image from its background. Second, the collected RGB values are device-dependent. Normally, the sensor receives the light and filters it to three channels (Red, Green, Blue) and their intensity values are determined by fruit samples based on the current illumination and the internal characteristics of the camera \citep{yam_simple_2004}. Third, accurate imaging needs homogeneous illumination in the field, which can be a very hard requirement to meet depending on the geographic location and weather conditions. 

\subsection{Volatiles}
\label{sec:volatiles}

Fruit types vary widely in aroma characteristics due to differences in the composition of the aromatic volatiles present in fruit aromas, which are ultimately determined by plant genetics \citep{references_1_v_prasanna_fruit_2007}. Fruit produces a wide range of volatile organic compounds that impact their characteristically distinct aromas and contribute to unique flavor characteristics. Fruit aroma and flavor characteristics are of key importance in determining consumer acceptance in commercial fruit markets based on individual preference. Previously, professional human graders have been used to judge fruit quality based on visual and aroma characteristics for selecting and evaluating fruit for ripeness at harvest and saleability in commercial fruit markets. The advent of electronic-nose (e-nose) devices has offered new alternative tools for grading fruit for ripeness. These instruments are very effective in discriminating complex mixtures of fruit volatiles, and thus are new promising efficient, and effective tools for classifying fruit based on its peculiar odor.

\subsection{Visible And Infrared Spectroscopy}
\label{sec:visible-infrared-spectroscopy}

When light hits the surface of the fruit, it can be absorbed, scattered, or re-emitted in a quantifiable manner. The amount of each of these quantities is determined by the physical properties and chemical constituents of fruit, and thus by its ripeness. Visible and Near InfraRed (VNIR) reflectance spectroscopy measures the reflected light in the range between 380 nm and 2500 nm. The light reflectance in this range is largely dependent on the light absorption by fruit samples. VNIR spectroscopy has been widely applied as a non-destructive and fast measurement method for multiple quality attributes of food ripening \citep{wang_fruit_2015}. More importantly, portable devices have been developed and used in the field for collecting spectral data \citep{makky_situ_2014}. 

\subsubsection{Spectral Indices}

Either visible or infrared spectra can be analyzed and related to different ripeness stages by using spectral indices. The whole wavelength scan, or values at key selected wavelengths, are used in regression models to correlate with specific attributes of a fruit item that are associated with ripeness. Using only one wavelength as an index for in-field assessment at a time is difficult because the values can be highly affected by the sensor, the environmental illumination, and the particle size \citep{huang_assessing_2021}. Spectral indices embed a degree of variety of the spectrum into a compact representation. They normally combine the surface reflectance at two or more wavelengths in order to indicate the relative abundance of a feature of interest. A number of key spectral indices have been calculated to describe the progressive change of peel pigment concentration during ripening. For example, the index of absorption (IAD) introduced by \cite{ziosi_new_2008} is a robust spectral index obtained by calculating the difference between the absorption at two wavelengths around the peak (670 nm and 720 nm). The IAD range was found to be similar across different growing seasons and showed a good correlation with ripeness for peaches \citep{lurie_maturity_2013}, apricots \citep{costa_establishment_2010} and nectarines \citep{bonora_combined_2015}. Selecting significant spectral indices is non-trivial. In a naive method, additional wavelengths are added, one by one, in order to strengthen the correlation with the principal component scores until none of the remaining wavelengths is significant. Another efficient automated method for identifying key wavelengths involves using a genetic algorithm, which employs natural selection and random mutations based on prediction accuracy. 

\subsubsection{Full Vs Selected Wavelengths}

The full wavelength or selected spectral indices can describe the change of peel pigment concentration during the ripening process and provide comparable values with the colorimetric method \citep{ferrer_changes_2005}. However, peel color is not always the only criterion for ripening assessment. The correlation between internal quality attributes related to ripening, such as firmness, was investigated with full or selected wavelengths from the VNIR spectra \citep{wang_fruit_2015}. For the full wavelengths, a common regression model is the Principal Component Regression (PCR), which uses Multiple Linear Regression (MLR) to correlate with the principal components scores extracted from the predictors \citep{mahesh_hyperspectral_2015}. This found practical application in some studies for fruit quality assessment (\textit{e.g.}, \cite{mahesh_hyperspectral_2015,rinnan_review_2009,gomez_non-destructive_2006}). Nevertheless, the variability of physical sample properties and the performance of the hardware can generate undesired results (\textit{e.g.}, light scattering, and random noise generated in the extracted spectra). These factors reduce the accuracy and robustness of the prediction models based on VNIR features \cite{mollazade_principles_2012}.

\subsubsection{Spectroscopic Methods}

Spectroscopic methods utilize longer wavelengths than colorimeter and visible imaging. However, similar to colorimeters, they are not likely to be applied as high-throughput ripening assessment tools due to their low spatial resolution. Spectroscopy has been used in assessing the ripeness of a large variety of fruit, and portable commercial spectrometers are available on the market for \textit{in-situ} data collection (\textit{e.g.}, \cite{saranwong_-tree_2003,camps_non-destructive_2009,makky_situ_2014}). Nevertheless, most of the studies have focused on the indoor, post-harvest assessment of fruit maturity. Inconsistent performances were observed for the models developed by spectra taken indoors and in the crop. Consequently, for outdoor ripeness assessment, it is necessary to build the prediction model with spectra taken in-field and understand the effect of environmental factors on the quality of spectra.

\subsection{Fluorescence}
\label{sec:fluorescence}

Fruit degreening (\textit{i.e.}, the loss of chlorophyll) is an effective indicator of fruit ripening. The measure of chlorophyll content of a fruit item can be detected using a fluorimetric sensor and proved to be correlated to its ripeness \citep{bodria_optical_2004}. One chlorophyll fluorimetric method measures the photochemical and non-photochemical processes with the illumination of actinic light \citep{royer_fluorescence_1995}. At the time being, fluorometers based on Pulse-Amplitude-Modulation (PAM) are commercially available. These use visible blue light to generate a fluorescence excitation and measure the minimum ($F_0$) and maximum ($F_m$) emitted fluorescence. The maximum quantum yield is calculated as $(F_m-F_0)/F_m$. This parameter was found to be negatively correlated with the ripening stage of, \textit{e.g.}, apples \citep{song_changes_1997}. This chlorophyll fluorimetric method is popular within laboratories, but difficult to apply in-field as the samples need to be dark-adapted. In the study by \cite{bodria_optical_2004}, they designed a fluorescence imaging system that measured the light emission at 690–740 nm with the excitation light in UV-blue and red regions. The performance of the system shows a good correlation between ripeness and fluorescence values, even though the hue of the skin color showed little change. The fruit samples used for measurement by this fluorescence imaging system were not dark-adapted, but the equipment has only been designed for laboratory use. A problem related to using fluorescence is thus the contamination by environmental factors, which limits the \textit{in-situ} application. In order to reduce the influence of environmental factors on the absolute fluorescence intensity at a single band, more studies were focused on understanding the fluorescence ratios using various light sources of defined wavelengths. This research led to the development of a handheld, multi-parametric fluorescence sensor: Multiplex® (Force-A, Orsay, France), which employs four LED light sources and three synchronized fluorescence detectors \citep{ghozlen_non-destructive_2010}. Nevertheless, current studies of fluorescence sensors are focused only on the analysis of specific parts of a fruit, which again limits their potential use for high throughput measurements in the field.

\subsection{Spectral Imaging}
\label{sec:spectral-imaging}

Spectral imaging consists in a type of imaging that uses multiple bands across the electromagnetic spectrum. Spectral data is collected using dedicated commercially available tools called spectrometers. However, several factors contributed to the limited use of spectrometers in the field and the lack of consumer applications. Although size and cost are the more obvious factors, others like stray light and reliability have contributed to making the challenge even more difficult. \cite{das_ultra-portable_2016} demonstrate a smartphone-based spectrometer design that is standalone and supported on a wireless platform. This device addresses the issues of size and cost of \textit{in-situ} data collection. The device can be used for rapid sorting in storage facilities for different varieties of fruit and to assess ripening. The portable nature of the device along with simplified data collection methods can have a huge advantage in gathering large sets of data that may be useful in building ML-based models. Smartphone-based fruit testing can be beneficial for several end-users. Currently, much of the testing of ripeness in apples and several other fruits is carried out in a destructive manner using mechanical tests. Optical tests are non-destructive and can assist farmers in determining optimum harvest times. All essential components of the smartphone-based spectrometer, like the light source, filters, micro-controller, and wireless circuits, have been assembled in a housing of reduced dimensions (\textit{i.e.}, 88mm×37mm×22mm), and the entire device weighs 48g. The resolution of the spectrometer is 15 nm, delivering accurate and repeatable measurements. The device has a dedicated app interface on smartphones to communicate, receive, plot, and analyze spectral data. The performance of the smartphone spectrometer is comparable to existing top spectrometers in terms of stability and wavelength resolution.

\subsubsection{Hyperspectral Imaging}

A well-studied form of spectral imaging is HyperSpectral Imaging (HSI), which generates a three-dimensional imaging cube with images at a range of continuous wavelengths. A single spectrum can be extracted from each individual pixel representing the absorption properties and textural information of fruit samples, which are correlated with their ripeness \citep{elmasry_image_2016}. Similarly to traditional visible imaging and spectroscopic methods, HSI is non-destructive and requires little sample preparation, but it has the added advantage that it can record both spatial and spectral information simultaneously \citep{mahesh_comparison_2015}. For the assessment of fruit ripeness, a common type of wavelength dispersion device is normally used, coupled with an imaging sensor for the HSI image acquisition. This is called \textit{line scanning}. A line scanning device has the imaging spectrograph dispersing the incident light into different wavelengths instantaneously between the visible and the near-infrared wavelength range (380–1700 nm). Line scanning HSI cameras scan the samples continuously in one direction, so they can be attached to moving platforms. Experiments have been made using tractors \citep{bauriegel_early_2011}, robots \citep{zhao_robust_2016} and Unmanned Aerial Vehicles (UAV) \citep{honkavaara_hyperspectral_2012}. Overall, HSI is a promising technique for fruit ripeness assessment. Nevertheless, the in-field application of this technique will need to overcome the challenges of handling the large data output and the calibration of variable light levels whilst in the crop.

\subsubsection{Multispectral Imaging}

MultiSpectral Imaging (MSI) is a form of HSI that collects data at specific wavelengths instead of scanning the whole wavelength range. This can be accomplished using a frame scanning imaging system with Liquid Crystal Tunable Filter (LCTF) coupled with a CCD or CMOS sensor. In \cite{lu_hyperspectral_2006} the authors use five wavelengths, based on previous studies, to correlate their scattering profiles with the ripeness of apples using an ANN. Another lower-cost MSI system uses a rotating filter wheel containing a few bandpass filters instead of LCTF. However, the tuning speed is lower than LCTF. This device has been used to predict the ripeness of peaches. With the best combination of four wavelengths, high correlation coefficients were achieved \citep{muhua_non-destructive_2007}. Similarly, \cite{liu_feasibility_2015} used MSI with 19 wavelengths to predict the ripeness of strawberries. PLS, SVM, and ANN were compared with the best correlation coefficient of 0.94. Among the techniques described above, MSI is the most promising for in-field measurement as it can record high-resolution images at selected key wavelengths for the prediction of specific quality attributes. Compared with HSI, MSI has a lower cost and is easier to convert into portable devices. A portable MSI device with four narrow-band light sources and four reflectance sensors of different wavelengths at 570, 670, 750, and 870 nm has been developed commercially \citep{lurie_maturity_2013}. This device was used to classify palm oil into different stages of ripeness using quadratic discriminant analysis and discriminant analysis with Mahalanobis distance classifiers, achieving a correct classification rate of $>85\%$.

\section{Data Preprocessing}
\label{sec:data-preprocessing}

In order to improve data analysis for the fruit classification task, a number of studies have applied different data preprocessing techniques to the spectra obtained before modeling \citep{rinnan_review_2009}. As a preprocessing tool, Savitzky–Golay (SG) is the most frequently used digital data smoothing filter \citep{olarewaju_non-destructive_2016,jha_non-destructive_2006,mcglone_internal_2003}. This applies the Linear Least Squares method to fit low-degree polynomial data. However, SG has contrasting effects on the performance of multivariate statistical models. For example, \cite{jha_modeling_2007} compared multiple preprocessing techniques and found that smoothing did not produce any improvement in comparison with other methods for the assessment of the ripeness in mangoes. Thus, they warn researchers and developers against the use of this technique.

Standard Normal Variate (SNV) \citep{sanchez_non-destructive_2012} and Multiple Scattering Correction (MSC) \citep{olarewaju_non-destructive_2016-1} are the two of the most frequently used techniques for photon scattering correction. MSC is used to eliminate the nonlinear scattering due to the non-uniform travel distance of light by linearising each spectrum to a reference spectrum (usually, the mean). Previous research has shown the similarity between SNV and MSC. For instance, it has been confirmed that the correlation coefficients were the same when assessing the sugar content of peaches using PLS models with SNV and MSC \citep{uwadaira_examination_2018}. SNV can, however, be applied to an individual spectrum without requiring a reference \citep{sanchez_non-destructive_2012}. In some studies, SNV was performed with de-trending, which was used to correct the baseline shift of spectra \citep{sanchez_non-destructive_2012}. Generating derivatives of spectra is a useful pre-processing technique to enhance subtle differences and reduce the effect of mirror reflection \citep{ragni_non-destructive_2012}. 

\section{Approaches To Classification}
\label{sec:approaches-to-classification}

\begin{figure}[t]
    \centering
    \includegraphics[width=0.49\textwidth]{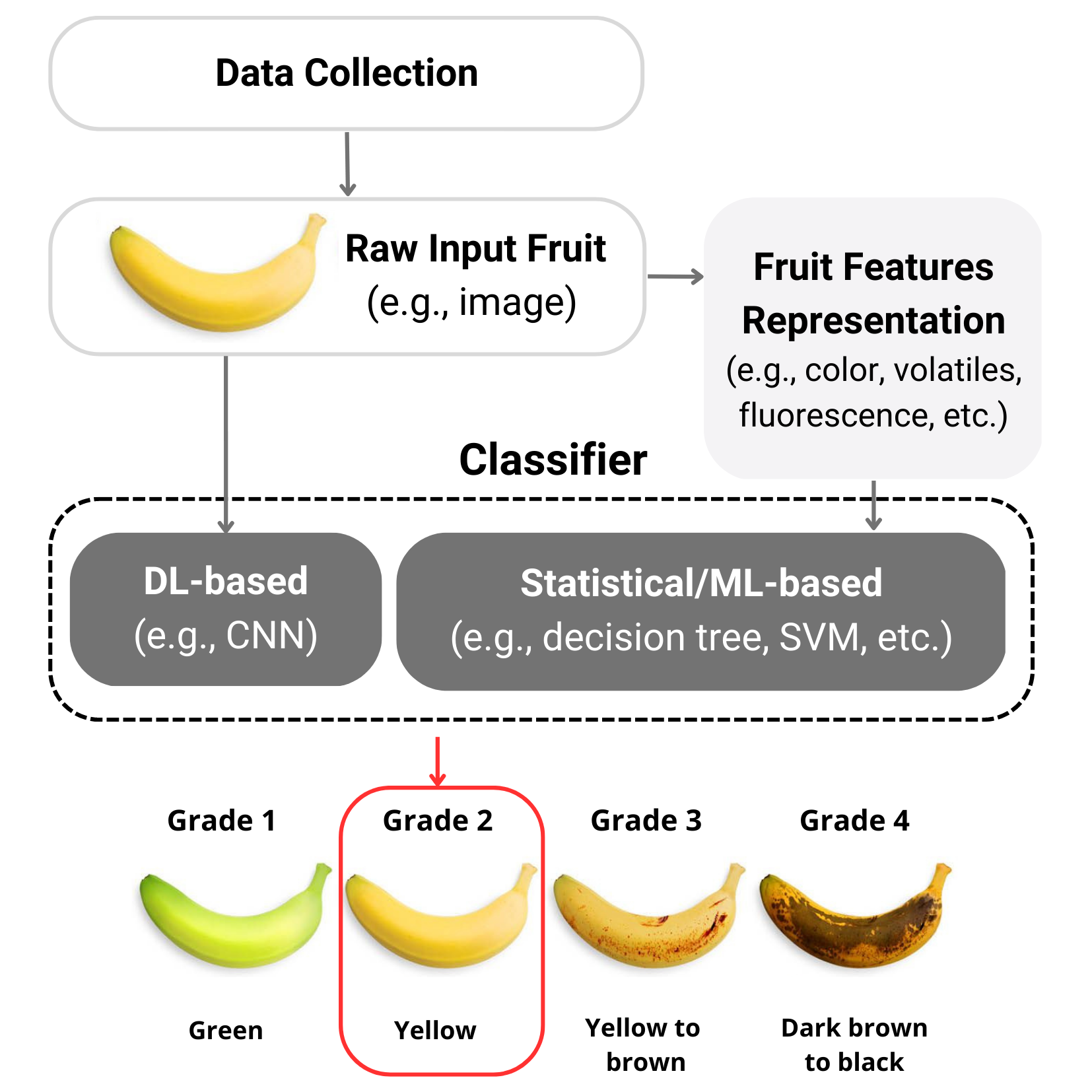}
    \caption{Example workflow for fruit ripeness classifiers.}
    \label{fig:classification-workflow}
\end{figure}

The task of fruit ripeness stage classification has been studied for decades. This has given birth to a vast literature on the topic. Our review is based on peer-reviewed papers published between 2014 and 2022. We deem this range of years appropriate for allowing a good trade-off between the depth and breadth of the survey. Among this corpus of literature, we identify three main groups of methods, that are: statistical, ML-based, and DL-based (regardless of the fruit type involved). Much of the seminal literature regarding fruit ripeness classification employs statistical models to address the task. At the same time, ML provides a toolbox of powerful algorithms that allow accurate classification performance. Some of these methods have been applied to the fruit ripeness classification task. However, in recent years, DL methods have taken over traditional techniques and helped to push state-of-the-art accuracy without the need to compute complex engineered features. Investigations were conducted exploring multiple different attributes of the fruit items (\textit{e.g.}, color, texture, smell). The next sections contrast statistical and ML-based methods with those employing DL strategies. Among the former group, we describe in detail what has been experimented on based on categories of feature representation (as previously introduced in §\ref{sec:feature-representations}). Fig \ref{fig:classification-workflow} depicts a standard workflow for a fruit ripening classification method, from the raw input to the output grade prediction. This highlights the core difference between DL-based classifiers, operating on raw data, and other types of classifiers, operating on rich feature representations.

\subsection{Statistical- and Machine-Learning-based}
\label{sec:statistical-ml}

\subsubsection{Colour}

One core attribute for fruit ripeness classification is color. A common standardized color space is the CIELAB color space, also referred to as L*a*b*. In a seminal work in this field, \cite{mendoza_application_2004} employ the CIELAB color space to enable the discrimination among seven different grades of banana ripeness previously labeled by experts in the field. They declare three objectives: (i) implement a standardized computer vision system to characterize quantitatively color changes during the ripening of bananas using the CIELAB color space, (ii) identify features of interest that can be related to ripening stages (\textit{e.g.}, development of brown spots and textural features of the images), and (iii) analyze the features extracted from images for their discriminatory power. They collected two proprietary datasets. The first was constituted of six hands of bananas at the ripening stage 1 (green) with thirteen or fifteen fingers per hand, stored at standard room temperature and humidity until used. Bananas were taken randomly from each hand. Their color changes and development of brown spots were measured daily over twelve days. The second dataset counts forty-nine bananas from a single batch, stored similarly, and visually selected by qualified banana industry workers according to seven ripening stages, seven samples per stage. In this case, the color, brown spots, and textural features were extracted from the banana images and employed for prediction. The background of pictures of bananas was removed at the preprocessing stage of the images using a threshold of 50 in the grey scale combined with an edge detection technique based on the Laplacian-of-Gauss (LoG) operator. The development of brown spots on the peel of bananas was evaluated from binarised images and quantified by two indexes: brown spots as a percentage of the total area (\%BSA) and the number of brown spots per $cm^2$ of surface ($NBS/cm^2$). Four textural features (homogeneity, contrast, correlation, and entropy) were extracted from images. Image texture was analyzed by studying the spatial dependence of pixel values represented by a co-occurrence matrix $P_d$ with entry $P_d(i, j)$ being the relative frequency or distance for two pixels d-pixels apart in direction  to have values i and j, respectively. The methods applied were simple regression, analysis of variance, and discriminant analysis with a 95\% confidence level. In spite of the inherent variability of banana samples, the proposed computer vision technique showed great potential to differentiate among ripening stages of bananas using simple features (L*, a*, b* bands, \%BSA, and contrast) extracted from the appearance of the peel. Using a basic discriminant analysis technique as the classification criterion, it was possible to identify forty-nine bananas in seven ripening stages with an accuracy of 98\%. Thus, the work by \cite{mendoza_application_2004} paves the ground for much further research in the field. Other classification methods were explored in the CIELAB color space. For example, \cite{olarewaju_non-destructive_2016} compared Multiple Linear Regression (MLR), Partial Least Squares (PLS), and Principal Component Regression (PCR) to predict the maturity stage of avocados. The former method proved to be the top performing on a self-collected proprietary dataset. 

The analysis of features performed by \cite{mendoza_application_2004} has helped develop statistical and ML models using a minimal set of selected features. However, note that in the same work images of bananas were processed including their background. On the contrary, more recent methods (\textit{e.g.}, \cite{ni_deep_2020,septiarini_automatic_2019}) use semantic segmentation to isolate the fruit items from their background. This design choice can have crucial repercussions as it relieves the risk that the classifier learns to discriminate based on the features of the context (\textit{i.e.}, the background) rather than the target item's features.

Further research was conducted on different color spaces. Moving on to the RGB color space, in \cite{goel_fuzzy_2015} they used the difference between R and B values to enhance the classification of the different tomato ripeness stages, reaching 94.3\% accuracy on a proprietary dataset of 250 images. Other statistical methods investigated include unsupervised classification, such as the K-means and Gustafson-Kessel algorithms. These have also successfully been applied to automatically separate bananas of different ripeness stages based on RGB values \citep{pardede_fruit_2019}. In later research, rather than only using the average RGB values, the histogram of each channel was used to find matches with predefined reference histograms for each ripeness group \citep{satpute_color_2016}.

The values of other color spaces such as Hue Saturation and Intensity (HSI) and Hue Saturation and Value (HSV), can be derived from RGB values and better represent human visual perception \citep{bakar_ripeness_2013}. The Hue is defined as the degree of similarity to some colors (usually, red, green, blue, and yellow) \citep{hartmann_htpheno_2011} whilst saturation is used to describe how colorful a stimulus is relative to its own brightness \citep{sural_segmentation_2003}. Fuzzy logic was successfully applied to group pineapples into three ripeness stages using values derived from HSI \citep{bakar_ripeness_2013}. Similarly, \cite{ukirade_color_2014} used HSV values as the input of an ANN model to classify tomatoes into four ripeness groups. \cite{el-bendary_using_2015} proposed a more sophisticated method for tomato ripeness classification. This employed the color histogram in the HSV space and the color moments (\textit{i.e.}, mean, standard deviation, and skewness), which measure the color distribution in an image as color features. Principal Component Analysis (PCA) was applied to extract the features for both Linear Discriminant Analysis (LDA) and SVM models with more than an 80\% correct classification rate for five ripening stages. Note that rather than only using one color space, color components from two color spaces can be combined, such as in the study by \cite{li_identifying_2014}. They used the RBH color space from outdoor color images of blueberries in combination with a KNN classifier. \cite{el-bendary_using_2015} propose a similar approach to classifying tomatoes in five ripeness stages utilizing color features. It consists of three phases: (i) pre-processing, (ii) feature extraction, and (iii) classification. During the pre-preprocessing phase, images are resized to 250x250 pixels, in order to reduce their color index, and the background of each image is removed using a background subtraction technique. Also, each image is converted from the RGB to the HSV color space. For the feature extraction phase, PCA was applied in order to generate a feature vector for each image in the dataset. Finally, for the classification phase, the proposed approach applied SVMs and LDA algorithms for the classification of ripeness stages. In particular, they use PCA in addition to SVMs and LDA algorithms for feature extraction and classification, respectively. Experiments have been conducted on a dataset of a total of 250 images. They showed a ripeness classification accuracy of 90.80\%, using the one-against-one multi-class SVM algorithm with a linear kernel function, a ripeness classification accuracy of 84.80\% using a one-against-all multi-class SVMs algorithm with a linear kernel function, and a ripeness classification accuracy of 84\% using LDA algorithm. Along the same lines, \cite{nambi_scientific_2015} describe the study carried out to classify the ripening stage of mangoes into different stages. They collected a proprietary dataset based on the physicochemical properties, external and internal color values, and textural characteristics that were measured throughout the ripening period of two Indian mango varieties. Experts identified five stages of ripeness for mangoes (\textit{i.e.}, unripe, early ripe, partially ripe, ripe, and over-ripe). They achieved promising results performing mango ripeness classification using PCA along with hierarchical clustering.

For a more comprehensive understanding of the problem, \cite{castro_classification_2019} evaluate the combinations of four ML techniques and three color spaces (RGB, HSV, and L*a*b*) with regard to their ability to classify cape gooseberry fruit. To this end, 925 samples were collected, and each fruit was manually classified into one of seven different classes according to its grade of ripeness. The classification of ripeness of cape gooseberry fruit was sensitive to both the color space and the classification technique used. The models based on the L*a*b* color space and the SVM classifier showed the highest f-measure regardless of the color space, and the PCA combination of color spaces improved the performance of the models at the expense of increased complexity.

\subsubsection{Volatiles}

The aroma of a given category of fruit is ultimately determined by plant genetics and serves as a major discriminant for its classification \citep{references_1_v_prasanna_fruit_2007}. Before the advent of electronic noses, human graders had used to judge fruit ripeness in a rather qualitative manner. E-nose devices have offered new alternative tools for grading fruit for ripeness. These instruments are very effective in discriminating complex mixtures of fruit volatiles, as new effective tools for more efficient fruit aroma analyses. The work by \cite{baietto_electronic-nose_2015} explores the current and potential utilization of electronic-nose devices (with specialized sensor arrays) in the fruit ripeness classification domain. E-noses contain a sensor array that evaluates all of the chemical constituents present in an aroma mixture (as a whole sample). Then, it converts the electronic output signals (via a transducer) from all of the sensors in the array and collectively assembles them to form a distinct digital pattern, sometimes referred to as an Electronic Aroma Signature Pattern (EASP). This is highly unique and specific to the particular gas mixtures being analyzed \citep{references_1_v_prasanna_fruit_2007}. In this way, the instrument output generates an aroma signature or smell print that can be used to identify the particular type and variety of fruit being analyzed. Even though different types of fruit share some aromatic characteristics, each fruit has a distinctive aroma. In a seminal work, \cite{llobet_non-destructive_1999} determined the state of ripeness of bananas by sensing the aromatic volatiles emitted by the fruit. For this purpose, they used an electronic artificial nose system in combination with a pattern-recognition engine. They found that their system led to a promising performance, providing 90\% accuracy in the classification of fresh bananas. Similarly, \cite{aghilinategh_detection_2020} use fruit odor to classify samples of berries into five classes. For classification, they employed ANN, PCA, and LDA analysis. The ANN achieved a precision of 100\% and 88.3\% for blackberry and white berries, respectively. Also, PCA analysis characterized 97\% and 93\% variance in the blackberry and white berry, respectively. The least correct classification for white berries was observed in the LDA method.

\subsubsection{Fluorescence}

As mentioned, a peculiarity of each ripeness stage for many varieties of fruit is the amount of chlorophyll present in an item. Initial work by \cite{li_optical_1997} developed an optical chlorophyll-sensing system to detect the chlorophyll content of bananas as the fruit ripens, showing a high correlation with other methods to assess the color of the peel (\textit{e.g.}, spectral analysis, instrumental analysis, and visual color matching). Common informative indices utilize fluorescence from anthocyanins (ANTH), flavonols (FLAV), and chlorophyll (CHL) to indicate fruit ripeness. In the study by \cite{li_optical_1997}, CHL showed a positive correlation with the ripeness of apples. The in-field assessment of CHL was also successfully applied to grapes, and the combination of CHL and ANTH can be used as a robust decision tool to predict ripeness as well \citep{li_advances_2018}. For tomatoes, all the indices were found to correlate well with the time shift in the tomato ripening process \citep{abdelhamid_nondestructive_2021}. The blue-to-red fluorescence ratio ($B_UV/RF_UV$) was measured as an effective parameter for the assessment of the ripeness of palm oil with rough skin, and when combined with the classification and regression tree method resulted in an overall correct classification rate of 90\% for three different ripeness stages \citep{hazir_oil_2012}. Finally, in \cite{das_ultra-portable_2016} Ultra-Violet (UV) fluorescence from Chlorophyll present in the skin was measured across various apple varieties during the ripening process and correlated with destructive firmness tests. A satisfactory agreement was observed between ripeness and fluorescence signals. This demonstration is a step toward the possible consumer, bio-sensing, and diagnostic applications that can be carried out in a rapid manner.

\subsubsection{Spectral Imaging}

Hyperspectral imaging is a data representation for fruit ripeness classification that proved to be valuable in a number of contexts. This can be handled in two different ways: (i) light scattering analysis and (ii) spectral analysis. Modified Lorentzian distribution, which correlates the data obtained with a predefined distribution curve by using a distribution function, can be used to describe the scattering profile and the fitting parameters that were used as the variables for a step-wise MLR model \citep{peng_analysis_2008}. The results of the study by \cite{qin_prediction_2009} suggest that spectral scattering from all wavelengths or selected wavelengths can provide more accurate predictions of apple ripeness than using secondary properties such as spectral absorption. Similar methods were also employed for the prediction of peach ripeness \citep{lu_hyperspectral_2006}. However, with MLR models different results were obtained when using two different cultivars (\textit{i.e.}, types of plants that people have bred for desired traits). \cite{mendoza_integrated_2011} combined both the spectral and image analysis techniques on scattering images, including discrete and continuous wavelet transformation decomposition, first-order statistics, Fourier analysis, co-occurrence matrix, and variogram analysis, but little improvements in the prediction of ripeness of apples were found. On the other hand, \cite{wang_model_2012} used two different feature selection methods: uninformative variable elimination \citep{centner_elimination_1996} and supervised affinity propagation \citep{zhu_wavelength_2013}. The output of two PLS models with two feature selection methods was combined as the input to an ANN model that gave a correlation coefficient of 0.83 with fruit ripeness.

The average spectra of the region of interest have also been modeled for the assessment of fruit ripeness. Key wavelengths have been selected, with different feature selection techniques. This operation was performed before modeling in order to reduce the redundancy of the whole spectral dataset. One of the widely used feature selection criteria is based on beta coefficients derived from PLS models. The PLS models measure how great an effect an independent variable has on the dependent variable. A comparison of the performance of different MLR models with wavelength selection based on beta coefficients and PLS models with full spectra as input showed that the final outcomes were similar for the ripeness grading of strawberries \citep{elmasry_hyperspectral_2007}. The same feature selection methods were also used by \cite{rajkumar_studies_2012} to predict the ripeness of bananas by MLR model and achieved a good correlation for ripeness. Another key wavelength selection method that can solve the collinearity problem is the successive projection algorithm, which iteratively adds wavelengths one by one until a specific number of wavelengths is achieved with a minimum redundant information content. This method has been used to select the feature wavelengths for input to the PLS model and a high correlation was found for predicting persimmon ripeness \citep{wei_ripeness_2014}. 

Near-Infrared Spectroscopy (NIRS) is another feature description method for fruit that was applied successfully to the prediction of the ripeness of many varieties of fruit. For instance, \cite{munawar_near_2019} evaluate the level of maturity of mangos using NIRS. The classification of mango ripeness was successfully achieved using second derivative pretreated spectra with an accuracy of more than 80\%. With a different approach, \cite{silalahi_using_2016} employed a genetic algorithm neural network for multi-class prediction of the ripeness grades of oil palm fresh fruit using NIRS spectral data. This data provided sufficient information about the compound structure of samples from the near-infrared light that passes through. The variables used in the GANN modeling process were the new variables obtained as a result of dimensional reduction from original NIRS spectral data using PCA. Three statistical measures (\textit{i.e.}, mean absolute error, root mean squared error, and the percentage of right classifications) were used to assess the adequacy of the model. This was precise enough to be used for the model calibration for fruit ripeness classification.

\subsection{Deep-Learning-based}
\label{dl}

\subsubsection{CNN And Pretrained Models}

In recent days, DL-based image processing is considered the state-of-the-art computer vision technique for image classification tasks. The introduction of CNNs made the feature engineering task way simpler as such networks are able to extract semantic features automatically through convolutions. Still, the visual classification of various stages of maturity of fruit is a challenging task as it may be hard to differentiate the visual features at different maturity stages. A recent study proposed to classify four different ripeness stages of bananas using a novel CNN architecture, which is compared with the state-of-the-art CNN model using transfer learning (\textit{i.e.}, VGG16 and ResNet50) \citep{saranya_banana_2022}. The proposed model performed comparably to the state-of-the-art but it is lightweight enough to be deployed on low-tier hardware. Further experiments were conducted involving transfer learning techniques using the VGG16 model. The proposed architecture uses VGG16 without the top layer. This was replaced by adding a Multilayer Perceptron (MLP) block for feature reduction and final classification. The MLP block contains a flattening layer, a dense layer, and regularizers. The output of the MLP block uses the softmax activation function. There are three regularizers that are considered in the MLP block, namely dropout, batch normalization, and regularizers kernels. The selected regularizers are intended to reduce overfitting. Based on the experimental results, the pretrained model seems to perform best. At the same time, the determination of the type of regularizers is very influential on system performance. The best performance was obtained on the MLP block that has a Dropout of 0.5 with increased accuracy reaching 18.42\%. The Batch Normalisation and the regularizer kernels performance increased the accuracy  by 10.52\% and 2.63\%, respectively. Ultimately, this study shows that the performance of DL strategies using transfer learning is always better than using ML with traditional feature extraction to determine fruit ripeness. 

Another discriminant characteristic of fruit that must be accounted for when grading their ripeness is whether the fruit is climacteric. \cite{wismadi_detecting_2020} focus on a non-climacteric fruit, that is the dragon fruit. This requires particular attention as it has to be harvested after it is ripened and cannot be ripened after harvesting using the hastening ripening process such as ethylene, carbide, CO2, etc. They put forth an application to identify the ripeness of the dragon fruit and the optimal harvest time using the RESNET 152 CNN-based model. The system they developed was trained using pictures of dragon fruit at different stages of maturity. The results they achieved were promising and more accurate compared to the VGG16/19.

\subsubsection{Detection-based Approaches}

It is common knowledge that DL algorithms work as well as the size and variety of their training set. A collection of relevant datasets for the task is summarized in Table \ref{tab:datasets}. The work of \cite{sa_deepfruits_2016} introduced the DeepFruits, a huge dataset for the task for fruit detection task, and a novel DL approach serving the same purpose. First, they explored the use of the FasterRCNN framework \citep{ren_towards_2015} for fruit detection and achieved impressive results. Furthermore, they demonstrated that such an approach could be rapidly trained and deployed with a small amount of training data (as few as 25 images). However, an aspect they did not explore was the potential for the system to perform not only fruit detection but also ripeness estimation. Inspired by DeepFruits, the work by \cite{halstead_fruit_2018} presents a robotic vision system that can accurately estimate the ripeness (as well as the quantity) of sweet pepper, a key horticultural crop. This system consists of three parts: (i) detection, (ii) ripeness estimation, and (iii) tracking. Similarly to the approach by \cite{sa_deepfruits_2016}, efficient detection is achieved using the FasterRCNN framework \citep{ren_towards_2015}. Ripeness is then estimated in the same framework by learning a parallel layer which experimentally shows superior performance than treating ripeness stages as extra classes in the traditional FRCNN framework. The evaluation of these two techniques outlines the slightly improved performance of the parallel layer: they achieve an F1 score of 77.3 using the parallel technique and 72.5 for the other best-scoring multi-class classifier. To track the crop, they present a vision-only tracking via detection approach, which uses the FRCNN with parallel layers as input. Being a vision-only solution, this approach is cheap to implement as it only requires a (possibly consumer) camera.

\begin{table}[h] 
    \footnotesize
    \begin{tabularx}{1.05\textwidth}{lclX} 
        \textbf{Name} & \textbf{Task} & \textbf{Sources} \\ 
        \hline 
        Fruit 360 & Classification & \cite{Fruit360, Fruit360dataset}\\ 
        
        Fruit recognition & Recognition & \cite{fruit_recognition, fruit_recognition_dataset} \\                
        ACFR Orchard Fruit & Detection & \cite{2016_multifruit, DeepFruitDetectionDataset} \\
        
        VegFruit & Classification & \cite{vegfru} \\

        MinneApple & Detection & \cite{MinneApple} \\

        Lemon & Quality assurance &  \cite{lemon_dataset} \\

        Grocery Store & Classification & \cite{GroceryStore} \\

        AppleScab FDs / LDs & Quality assurance & \cite{applescab_article, AppleScabFDs, AppleScabLDs}  \\

        Pistachio & Detection & \cite{pistachio_image, pistachio_image_dataset}\\
        
        Date fruit & Detection & \cite{date_image, date_image_dataset} \\

    \end{tabularx} 
\caption{Public datasets used within fruit-related applications.} 
\label{tab:datasets}
\end{table}

Building a more complex system, the goal of the study proposed by \cite{ni_deep_2020} is to develop a data processing pipeline to count berries, measure their maturity, and evaluate the compactness (cluster tightness) automatically. Their strategy employs a DL-based image segmentation method to extract an individual or groups of berries from the background. The target consists of four southern high-bush blueberry cultivars (\textit{i.e.}, Emerald, Farthing, Meadowlark, and Star). A Mask R-CNN \citep{he_mask_2017} model was trained and tested to detect and segment individual blueberries with respect to their maturity stage. The mean average precision was 71.6\% and the corresponding mask accuracy was 90.4\%. The analysis of the traits collected from the four cultivars highlighted useful information for the producers. For example, it indicated that the ‘Star’ variety had the fewest berries per cluster, ‘Farthing’ had the least mature fruit in mid-April and the most compact clusters, while ‘Meadowlark’ had the loosest clusters. Thus, there is proof that the DL image segmentation technique developed in this study is efficient for detecting and segmenting blueberry fruit, extracting traits of interest related to machine harvestability, and monitoring blueberry fruit development.

\subsubsection{Combination Of Neural Models And Engineered Features}

The above-mentioned techniques were tested against a variety of fruit types at the same time. One particularly relevant crop worldwide is represented by bananas. The effectiveness and fast classification of the maturity stage of bananas are the most decisive factors in determining their quality. This is one of the reasons that motivate the necessity to design and implement image processing tools for the correct ripening stage classification of the different incoming banana bunches. In \cite{mazen_ripeness_2019} an automatic computer vision system is proposed to identify the ripening stages of bananas. First, a four-class homemade database is prepared. Second, an ANN-based framework that uses color, the development of brown spots, and Tamura statistical texture features \citep{tamura_textural_1978} are employed to grade the ripening stages of bananas. The performance of the proposed system is compared with various other traditional ML methods (\textit{e.g.}, SVM, naive Bayes, KNN, decision trees, and discriminant analysis). The results reveal that the proposed ANN-based system has the highest overall classification rate (97.75\%) among the considered techniques.

Similarly to the research by \cite{mazen_ripeness_2019}, other works specialized in grading one particular type of fruit. For instance, \cite{miraei_ashtiani_detection_2021} analyze, evaluate and classify mulberry fruit according to their stage of ripeness. Three stages were considered: unripe, ripe, and overripe. To collect their proprietary dataset, a total of 577 mulberries were graded by an expert and the corresponding images were captured by an imaging system. Then, individual mulberries were segmented and the geometrical properties, color, and texture characteristics of each item were extracted using two feature reduction methods. One feature reduction method is the Correlation-based Feature Selection subset (CFS) and the other is the Consistency subset (CONS). Finally, an ANN and SVM were applied to classify the mulberry fruit. The ANN classification paired with the CFS subset feature extraction method resulted in an accuracy of 100\%, 100\%, and 99.1\% over the three classes, respectively. Also, the same couple led to the least mean square error values. The ANN structure with the CONS subset feature extraction method still resulted in an acceptable model with a corresponding accuracy of 100\%, 98.9\%, and 98.3\%. In general, the machine vision system combined with the ANN and SVM algorithms successfully classified mulberries based on maturity.

Again using CNNs, \cite{liming_automated_2010-1} discuss an automatic vision-based system for sorting and analyzing strawberries. They proposed an automated system to predict the ripeness level of strawberry fruit, using a very simple CNN architecture. For the success of classification, appropriate features must be extracted. Surface color, size, and shape are necessary features for classification. The surface color of strawberry fruit determines its ripeness level. The CNN proved to extract color, size, and shape features from strawberry surfaces and achieves as high as 91.6\% accuracy.

On a different note, \cite{gao_real-time_2020} estimate the ripeness of strawberries using the HSI system, both in the field and in laboratory conditions. HSI data was collected for strawberries at early ripe and ripe stages, covering wavelength ranges from 370 to 1015 nm. Spectral  feature wavelengths were selected using the sequential feature selection algorithm. Two wavelengths were selected for the field (530 and 604 nm) and laboratory (528 and 715 nm) samples, respectively. Then, the reliability of such spectral features was validated based on an SVM classifier. The performance of the SVM classification models showed good results based on the receiver operating characteristic values for samples under both field and laboratory conditions (higher than 0.95). Additionally, a CNN was used to extract spatial features from the spectral wavelength and the first together with the three principal components for laboratory samples. A pretrained AlexNet CNN was used to classify the early ripe and ripe strawberry samples, which obtained an accuracy of 98.6\% for the test dataset.

Adding to the discussion on classification based on color features, \cite{rivero_mesa_non-invasive_2021} propose a non-invasive automated system for export-quality banana ripeness tiers. This system combines numerous features, including RGB values and hyperspectral imaging, put together by DL techniques. The multi-input model achieved an excellent overall accuracy of 98.45\% using only a minimal number of samples compared to other methods in the literature. Moreover, the model was able to incorporate both the external and internal properties of the fruit. The size of the bananas was used as a feature for ripeness grade classification as well as other morphological features using RGB imaging. Reflectance values of the fruit were also used and obtained through hyperspectral imaging. These proved to offer valuable information and have shown a high correlation with the internal features. This study highlighted the combined strengths of RGB and hyperspectral imaging in grading bananas, which may serve as a paradigm for grading other horticultural crops.

A similar system was developed by \cite{garillos-manliguez_multimodal_2021} to estimate six maturity stages of papaya fruit whilst suggesting a novel nondestructive and multi-modal classification method using deep CNNs. The models they designed estimated fruit maturity by feature concatenation of data acquired from two imaging modes: visible-light and hyperspectral imaging systems. Morphological changes in the sample fruit can be measured with RGB images. At the same time, the spectral signatures, which provide high sensitivity and correlation with the internal properties of the fruit, can be extracted from hyperspectral images with a wavelength range between 400 nm and 900 nm. In the study, some of the most popular pretrained CNN architectures (\textit{i.e.}, AlexNet, VGG16, VGG19, ResNet50, ResNeXt50, MobileNet, and MobileNetV2) were investigated to utilize multi-modal data cubes composed of RGB and hyperspectral data for sensitivity analyses. Such multi-modal variants could achieve up to 0.90 F1 scores for the six-way classification. These results confirm the above-mentioned findings indicating that multi-modal DL architectures and multi-modal imaging have great potential for real-time in-field fruit ripeness estimation.

\subsubsection{Classification In Orchard Environment}

Specializing in yet another fruit crop, \cite{altaheri_date_2019} fill the gap in research in the area of machine vision for date fruits in an orchard environment. In their work, they propose an efficient machine vision framework for date fruit harvesting robots. The framework consists of three classification models used to classify date fruit images in real-time according to their type, maturity, and harvesting decision. In the classification models, deep CNNs are utilized with transfer learning and fine-tuning on pre-trained models. To build a robust vision system, they create a rich image dataset of date fruit bunches in an orchard that consists of more than eight-thousand images of five date types in different pre-maturity and maturity stages. The dataset has a large degree of variations that reflects the challenges in the date orchard environment including variations in angles, scales, illumination conditions, and date bunches covered by bags. The proposed date fruit classification models achieve accuracy scores as high as 99.01\%, 97.25\%, and 98.59\% for the corresponding ripeness classes. Furthermore, they report times of 20.6, 20.7, and 35.9 milliseconds for the type, maturity, and harvesting decision classification tasks, respectively. Such values make the proposed system viable for real-time use.

\subsubsection{Real-time Assessment Of Ripeness}

The research by \cite{suharjito_oil_2021} addresses the problem of oil palm ripeness classification. Though many techniques have been explored in the literature, most of the methods require devices with high computational resources that could not be implemented in mobile applications. To overcome this problem, the authors focus on creating a mobile application to classify the ripeness levels of oil palm using a lightweight CNN. They implemented ImageNet transfer learning on four lightweight CNN models with a novel data augmentation method named “9-angle crop”, which can be further optimized using post-training quantization. Transfer learning with three unfrozen convolutional blocks and 9 angle crops successfully increased the classification accuracy on MobileNetV1. However, EfficientNetB0 performed best with an accuracy of 0.89. Float16 quantization also proved to be the most suitable post-training quantization method for this model, halving the size of EfficientNetB0 with the least increase in image classification time and an accuracy drop of only 0.005.

\subsubsection{Attention-based Approaches}

A recent breakthrough in DL has been the introduction of attention models \citep{bahdanau_neural_2015}, which found application, among others, in fruit ripeness classification. Attention allows learning a weighted sum of embeddings of the input tokens that can then be manipulated in various ways (\textit{e.g.}, for input classification). The study by \cite{herman_oil_2020} led to the development of a model for oil palm ripeness prediction using a residual-based attention mechanism ANN (ResAttDenseNet) that could recognize the small detail differences between images. The proprietary dataset used for conducting their experiments consists of four-hundred images including seven levels of ripeness. As a result, they showed that the proposed model could improve the F1 Score by 1.1\% compared to the highest F1 score from other basic DL and traditional ML models compared in the study.

\section{Prediction Of Optimal Harvest Time}
\label{sec:optimal-harvest}

Global food security for the increasing world population not only requires increased sustainable production of food but also process optimization. This means that a significant reduction in pre and post-harvest waste is imperative. The timing of when some fruit is harvested is critical for reducing waste along the supply chain and increasing fruit quality for consumers. The early in-field assessment of fruit ripeness and prediction of the harvest date and yield by non-destructive technologies have the potential to revolutionize farming practices and enable the consumer to eat the tastiest and freshest fruit possible. As discussed above, a variety of non-destructive techniques have been applied to estimate the ripeness or maturity but not all of them are applicable for \textit{in situ} (field or glasshouse) assessment. Non-destructive methods are very promising for in-field ripeness assessment, but the most critical question is how to link such assessment to predict yield and the optimal harvest date. This is highly challenging and made complicated by ripeness variability within and between plants.

Environmental factors such as temperature, light levels, humidity, etc. significantly influence the development of crops as well, and it is essential to incorporate the predictions of these important environmental factors in the determination of the optimal harvest date \citep{teng_study_2012}. Several crop models have been developed since the 1960s based on the work in \cite{loomis_maximum_1963} with the input of environmental factors both for in-field and greenhouse prediction. Such models are difficult to use due to the number of input variables (\textit{e.g.}, \citep{qiu_determining_2016}). In the research by \cite{qiu_determining_2016}, they investigated the dominant environmental factors in greenhouses for tomato growth and it was found that temperature, humidity, and photosynthesis active radiation interact according to a positive or negative correlation to crop growth. The influence of temperature has also been reported in a number of studies such as for tomatoes \citep{teng_study_2012}, grapes \citep{tomana_effect_1979}, apple \citep{yamada_effect_1994}, mangoes \citep{medlicott_effects_1986}, and more. The overall color change of the crop at different constant temperatures was studied, but the variation of the temperature within each day was not considered \citep{shewfelt_prediction_1988}. \cite{munoz_prediction_2012} developed a time series regression model for the prediction of the harvest date of blueberries. The minimum and maximum daily temperatures from the weather forecast for two weeks ahead were used as input for the model. This method was closer to a real application and potentially could be paired with non-destructive techniques to determine the current ripening stage of the crop.

Interestingly, it was found that environmental temperature affects not only fruit growth but also the near-infrared reflectance spectrum in a non-linear way \citep{ma_nondestructive_2007}. \cite{kawano_development_1995} compensated for the surface temperature effect by developing a combined MLR model, which covered a variety of temperatures ranging between 21 and 31 Celsius. On the other hand, \cite{peirs_temperature_2003} compared a global calibration model that covers a wide temperature range and calibration models for each temperature range. Both methods performed well for the ripeness prediction of apples, but for practical purposes, the global calibration model was preferred.

Finally, \cite{yang_remote_2011} recorded the HSI spectra of tomatoes at different growing stages, and the PLS model was applied to predict the growing stage of a target crop with the best correlation coefficient being 0.89. It was also observed that the key wavelengths were in the visible and infrared regions (400–2100 nm) \citep{yang_remote_2011}. A similar method was employed to predict the number of days until the commercial harvest of apple \citep{peirs_prediction_2001}. The calibration model was built with eight cultivars and a good correlation (R2 = 0.93) was found for the spectral range between 380 nm and 2000 nm. 

\section{Perspectives}
\label{sec:perspectives}

Despite the CNN models proving to work particularly effectively and efficiently for fruit ripeness classification, there is still room for improvement. First, the accuracy of the task might be boosted by using new state-of-the-art models such as Transformers \citep{vaswani_attention_2017}. These were initially devised for neural machine translation in the field of Natural Language Processing (NLP) but showed outstanding results in many other text-related tasks (\textit{e.g.}, \citep{khandelwal_sample_2019,sun_how_2019}). More recently, Transformer-based architectures were employed with little tuning to visual tasks, both image- \citep{han_survey_2023} and video-based \citep{neimark_video_2021}. Investigating the use of Transformers in the domain of fruit ripeness classification from a visual perspective could improve the accuracy and yield more robust models. Furthermore, the visual Transformer could be mixed with CNNs in different ways, the simplest interaction being late averaging of the output. On the other hand, insisting on the CNNs path would require refining their optimization, \textit{i.e.}, finding the optimal number of layers and filters for the specific problem to be solved, as well as determining the parameters and hyperparameters of the model. This remains a relevant problem commonly solved by trial and error until the best configuration is achieved, which is especially time-consuming for very deep models. Also, we warn against a gap in the literature on fruit ripeness classification when it comes to measuring inference time. Low inference time is a hard requirement for developing real-time applications, which are required to perform \textit{in situ} inference.

An obvious limitation to both the CNN- and Transformer-based approaches is the size of the available datasets. To solve a task using DL methods the dataset must be sufficiently large and well-labeled to address underfitting and overfitting problems. Therefore, the process of preparing the dataset is one of the activities that require more time and effort. Although there is a wide variety of publicly available datasets, not all researchers release their data. For this reason, the reproducibility of some studies is not entirely guaranteed. In addition, the datasets are often collected with a very specific task in mind, which limits the inference potential of the trained model. It is imperative that further studies release their code and datasets so that new models can be tested against a larger number of benchmarks to get a better overview of the challenges and achievements of the task.

On another note, developing AI-based models for food quality assessment applications means working in a high stake decision-making environment. Thus, one may want to investigate and compare the interpretability of the models used for fruit image processing (be they traditional ML models, CNNs, or Transformers) to provide explanations for both model developers and end users. In the first stage, one could aim for data debugging. This means analyzing the distribution of the data points and their features and searching for discrepancies. For example, a trivial check would require assessing whether the dataset is balanced or not. In the latter case, several methods can be used to make it balanced (\textit{e.g.}, \citep{chawla_smote_2002}). Second, one could statistically analyze the output distribution of the models, looking, for instance, for outliers. The output of this analysis itself could tell much about what the model has been learning and whether this is correct. On the other hand, it might be hard to make strong inferences based on it (\textit{e.g.}, justify the presence of outliers). Finally, sophisticated interpretability techniques exist for off-the-shelf usage (see \cite{carvalho_machine_2019} for a survey). Albeit with different levels of reliability, these may provide interesting information about black-box models.

\section{Conclusions}
\label{sec:conclusions}

This review offers a wide panoramic of the fruit ripeness classification task. More specifically, we offer a formal statement of the problem and a summary of the biological processes involved in fruit ripening. Then, we discuss the different types of descriptors that can be used to represent a fruit item: color, light spectrum, fluorescence, and spectral imaging. The featurized items can thus be processed by either statistical, ML, or DL models. When using former types it is important to select and possibly generate the right features, \textit{i.e.}, the most discriminative. On the other hand, DL models do not require feature engineering, which is a labor-intensive and error-prone process. Yet, they still achieve state-of-the-art accuracy for a variety of fruit ripeness classification tasks. Thus, it seems that DL models, especially pretrained and then fine-tuned models, are the most promising approach to fruit ripeness classification. However, we warn against their inner opacity: it is hard for humans to understand how a DL model made a specific prediction. Considering that ripeness is bound to food quality and safety, it is of the utmost importance to consider a trade-off between accuracy and interpretability. Hence, future research directions may want to investigate models that are more transparent. In particular, Transformer models for images are an emerging field that promises higher accuracy and higher interpretability by allowing researchers to examine the attention heads that compose the model. Even if some work debunked the use of attention for producing explanations \citep{jain_attention_2019,wiegreffe_attention_2019,serrano_is_2019}, whether it carries some intelligible information is still an open question in the eXplainable AI (XAI) community.

\section{Acknowledgments}
This paper was funded by Veneto Agricoltura within the scope of the project "Guaranteeing the continuity of the agri-food chain: the digitization of wholesale markets".

\bibliographystyle{unsrt}
\bibliography{paper}

\clearpage

\end{document}